\documentclass[
]{ceurart}

\usepackage{listings}
\usepackage{graphicx}
\usepackage{subcaption}
\usepackage[utf8]{inputenc} 
\usepackage[T2A]{fontenc}    
\usepackage[russian,english]{babel}
\usepackage{CJKutf8}
\usepackage{microtype} 
\usepackage{array} 
\usepackage{booktabs}
\usepackage{tabularx}
\usepackage{makecell}
\usepackage{tikz}
\usetikzlibrary{shapes.geometric, arrows}

\tikzstyle{startstop} = [rectangle, rounded corners, minimum width=2.5cm, minimum height=1cm, text centered, draw=black, fill=red!30]
\tikzstyle{process} = [rectangle, rounded corners, minimum width=2.5cm, minimum height=1cm, text centered, draw=black, fill=orange!30]
\tikzstyle{decision} = [diamond, minimum width=2.5cm, minimum height=1cm, text centered, draw=black, fill=green!30]
\tikzstyle{arrow} = [thick,->,>=stealth]
\tikzstyle{altprocess} = [rectangle, rounded corners, minimum width=2.5cm, minimum height=1cm, text centered, draw=black, fill=blue!30]

\begin{document}

\copyrightyear{2025}
\copyrightclause{Creative Commons License Attribution 4.0
  International (CC BY 4.0).}

\title{Evaluating Named Entity Recognition Models for Russian Cultural News Texts: From BERT to LLM}

\author[1]{Maria Levchenko}[%
orcid=0000-0002-0877-7063,
email=maria.levchenko@studio.unibo.it,
url=https://mary-lev.github.io/,
]
\address{University of Bologna}

\begin{abstract}
This paper addresses the challenge of Named Entity Recognition (NER) for person names within the specialized domain of Russian news texts concerning cultural events. The study utilizes the unique SPbLitGuide dataset, a collection of event announcements from Saint Petersburg spanning 1999 to 2019. A comparative evaluation of diverse NER models is presented, encompassing established transformer-based architectures such as DeepPavlov, RoBERTa, and SpaCy, alongside recent Large Language Models (LLMs) including GPT-3.5, GPT-4, and GPT-4o. Key findings highlight the superior performance of GPT-4o when provided with specific prompting for JSON output, achieving an F1 score of 0.93. Furthermore, GPT-4 demonstrated the highest precision at 0.99. The research contributes to a deeper understanding of current NER model capabilities and limitations when applied to morphologically rich languages like Russian within the cultural heritage domain, offering insights for researchers and practitioners. Follow-up evaluation with GPT-4.1 (April 2025) achieves F1=0.94 for both simple and structured prompts, demonstrating rapid progress across model families and simplified deployment requirements.
\end{abstract}

\begin{keywords}
  Named Entity Recognition \sep
transformer‐based models \sep
Large Language Models \sep
Russian cultural news \sep
benchmark evaluation \sep
Russian NLP
\end{keywords}

\conference{}
\maketitle
\section{Introduction}

\subsection{Background and Motivation}

Named Entity Recognition (NER) stands as a foundational task in Natural Language Processing (NLP), pivotal for extracting structured information from unstructured text. Its significance is increasingly recognized in specialized domains such as cultural studies, where the identification of entities like person names is crucial for analyzing historical trends, mapping intellectual networks, and constructing narratives. This evaluation aims to select an appropriate NER model specifically for analyzing Russian news texts pertaining to literary and cultural events and academic seminars.

The task is particularly demanding for morphologically rich languages like Russian. Extracting personal names from event descriptions, for instance, proves significantly more challenging due to the language's complex inflectional system, including multiple grammatical forms and cases, compounded by the unique stylistic characteristics of the news genre in the cultural domain. The advent of Large Language Models (LLMs), such as GPT-4 introduced by OpenAI in 2022, has marked a significant advancement in NLP, offering remarkable capabilities for various information extraction tasks, including NER. However, despite these advancements, challenges persist in ensuring the efficient, effective, and sustainable use of such models, especially in specialized fields. Issues like maintaining context, addressing biases, ensuring accuracy, understanding nuanced language, and managing computational resources necessitate ongoing research and optimization. This study explores strategies to harness the potential of LLMs for NER in the cultural domain while mitigating these limitations.

The application of advanced NLP techniques to cultural and historical texts in non-English languages is an area where the general capabilities of models like GPT-4 require specific validation. While these models demonstrate broad understanding, their efficacy in niche domains, particularly those involving complex linguistic features and specialized vocabularies, is not guaranteed out-of-the-box. This research, therefore, seeks to bridge the gap between the general advancements in LLMs and the concrete, domain-specific requirements of humanities research, particularly for Russian language materials.

\subsection{Problem Statement}

The core research problem is the pressing need for an effective and accurate NER model capable of identifying person names within the SPbLitGuide dataset. This dataset comprises Russian news texts detailing cultural events in Saint Petersburg from 1999 to 2019. The intended application of this dataset, once names are successfully recognized and matched, is to facilitate an in-depth analysis of the cultural dynamics and literary landscape of St. Petersburg over two decades. This includes tracking key figures, their connections within the literary community, mapping event locations, and identifying trends that have shaped the city's cultural scene. Given that the dataset contains over 15,000 events with more than 10,000 unique individual names mentioned in event descriptions, automating the NER process is a critical requirement for enabling such large-scale analysis. The inherent complexities of the Russian language and the specific genre of cultural news make off-the-shelf solutions potentially inadequate, necessitating a rigorous comparative evaluation to find a model that can handle these challenges effectively.

\subsection{Contributions}

This paper offers several key contributions to the field:
\begin{enumerate}
    \item A comprehensive evaluation of a diverse set of NER models on a unique, publicly available Russian cultural news dataset. The models range from established transformer architectures (DeepPavlov, RoBERTa, SpaCy) to state-of-the-art LLMs (GPT-3.5, GPT-4, GPT-4o, GPT-4.1).
    \item An analysis of model performance focusing specifically on ``PERSON'' entities, utilizing standard metrics: precision, recall, and F1-score.
    \item Insights into the effectiveness of different prompting strategies for LLMs in this NER context, particularly for the GPT-4o model, where variations in API requests and output format instructions were tested.
    \item Identification of the best-performing models tailored to specific requirements, such as achieving the highest precision or the highest recall, within this particular domain.
    \item Adherence to reproducible evaluation principles, as advocated by Biderman et al. (2024), through the sharing of data and model outputs to ensure transparency and facilitate further research.
\end{enumerate}

\section{Related Work}

\subsection{Evolution of NER Techniques}

The methodology for Named Entity Recognition has undergone a significant transformation. As surveyed by Ehrmann et al. (2023), NER approaches evolved from rule-based systems prevalent around 2002, transitioned to traditional machine learning techniques like Conditional Random Fields (CRF), and subsequently adopted deep learning methods. Between 2018 and 2020, models based on Bidirectional Long Short-Term Memory (BiLSTM) with CRF layers, and eventually BERT-like architectures, became prominent. This progression reflects a continuous pursuit of improved accuracy and efficiency in NER across diverse domains and languages. Each paradigm shift aimed to better capture contextual information and handle linguistic complexities, which is particularly relevant for the type of challenging dataset examined in the current study. The comparison of BERT-based models with the latest generation LLMs in this work is a natural continuation of this evolutionary trend, testing whether these newer models offer substantial advantages for complex Russian cultural texts.

\subsection{NER in Historical and Cultural Domains}

The application of NER to historical documents and cultural heritage texts has garnered increasing attention. Datasets such as NewsEye and HIPE-2020 serve as benchmarks for historical newspaper data, complemented by classical datasets like ajmc and CoNLL. Other datasets covering news, medical, social media, and reviews include BioNLP2004, WNUT2017, MIT-Movie, MIT-Restaurant, and BC5CDR, providing a broad basis for NER model evaluation.

Specific studies have advanced NER in these contexts. For instance, Boros et al. (2020) introduced the L3i NERC-EL model for multilingual NER, employing a hierarchical, multitask learning approach with a fine-tuned BERT encoder and additional Transformer layers. This model demonstrated strong performance on the HIPE historical dataset for English, French, and German. Todorov and Colavizza (2020) presented a modular embedding layer combined with a BiLSTM-CRF model for NER on historical corpora, finding that architectural simplifications could still yield effective results. These efforts highlight the ongoing adaptation of NER techniques to the unique characteristics of historical and cultural texts.

\subsection{Large Language Models for NER}

The investigation into using LLMs for NER tasks began gaining traction around 2023, with studies focusing on domains like cultural heritage and clinical text. González-Gallardo et al. (2023) evaluated ChatGPT's zero-shot capability to extract named entities from historical documents. Their findings revealed that, despite its power, ChatGPT yielded relatively poor results compared to traditional transformer-based models like Stacked and Temporal NERC. Factors contributing to this included inconsistent entity annotation guidelines, entity complexity, multilingualism, and the specificity of prompting. A significant limitation was also the unavailability of historical newspaper datasets to the web scrapers OpenAI used for training data.

In the clinical domain, Naguiba et al. (2024) found that masked language models outperformed LLM prompting for few-shot clinical entity recognition in three languages. Hu et al. (2024) explored prompt engineering to improve LLMs for clinical NER. Peng et al. (2024) proposed MetaIE, a method to distill a meta-model from an LLM for various information extraction tasks. These studies underscore that while LLMs show considerable promise, applying them to specialised Named Entity Recognition (NER) tasks, particularly for non-English historical or cultural content, is still an experimental process.

\subsection{Evaluation Practices in NLP}

Ensuring reproducibility and transparency in NLP research is key. Biderman et al. (2024) outlined principles for the reproducible evaluation of language models, including sharing code and prompts, providing model outputs, performing qualitative analysis, and measuring and reporting uncertainties. This study endeavors to adhere to these principles by making data and model outputs available, thereby contributing to verifiable research practices.

\section{Dataset and Task Definition}

\subsection{The SPbLitGuide Dataset}

The dataset employed in this study was created by parsing and cleaning raw text from electronic newsletters disseminated by the SPbLitGuide (Saint Petersburg Literary Guide) project between 1999 and 2019. These newsletters chronicled upcoming cultural events in Saint Petersburg, offering a unique source of information for analyzing the city's literary and cultural landscape over two decades.

The cleaned and structured dataset has been published on Zenodo \citep{levchenko2024}. It comprises 15,012 records of events, each detailed with attributes such as event ID, event description, date/time, location, address, and geographic coordinates (latitude/longitude). While parsing and structuring geodata like addresses and geocodes was relatively straightforward, the extraction of personal names from event descriptions posed a considerable challenge. The primary goal for this dataset is to enable an in-depth analysis of St. Petersburg's cultural dynamics and literary landscape from 1999 to 2019. This involves tracking key figures and their connections, mapping event locations, and identifying trends that have influenced the city's cultural evolution. The public availability of this dataset, along with a detailed characterization of its complexities, represents a significant contribution, offering a new benchmark resource for Russian NER, especially within the under-resourced cultural domain.

Table \ref{tab:dataset_stats} provides a summary of the dataset statistics.

\begin{table}[h]
\centering
\caption{SPbLitGuide Dataset Statistics}
\label{tab:dataset_stats}
\begin{tabular}{ll}
\toprule
Feature & Value \\
\midrule
Total Records & 15,012 \\
Time Span & 1999-2019 \\
Source & SPbLitGuide Electronic Newsletters \\
Annotated Sample Size & 1,000 records \\
Annotated ``PERSON'' Entities & 5,611 \\
Zenodo DOI & 10.5281/zenodo.13753154 \\
\bottomrule
\end{tabular}
\end{table}

\subsection{Annotation Process and Sample}

For evaluation purposes, a sample of 1,000 records was randomly selected from the entire dataset of 15,012 records, based on event date and event description length. This sample was manually annotated using the Doccano text annotation framework. The annotation process yielded a JSONL file containing text with associated labels, where each label has a start and end index indicating its position within the text.

Crucially, only the ``PERSON'' label was used during annotation, recognizing that named entities can appear multiple times within a single text. In this sample dataset of 1,000 texts, a total of 5,611 ``PERSON'' labels were identified. This focused approach was chosen in order to investigate the accuracy and efficiency of person name extraction, on which subsequent literary network analysis is based. By concentrating on individual names and excluding names embedded within organizational titles or works of art, this method aims to ensure that the dataset accurately reflects real participants, minimizing ambiguity or misclassification caused by different entity types While this focused annotation simplifies the labeling task, it simultaneously increases the complexity for the NER models. These models must perform fine-grained disambiguation, distinguishing person names from other name-like strings that might appear in titles of organizations or artistic works but are not the target entity type.

\subsection{NER Task Definition}

The NER task is defined as the identification and extraction of all mentions of real person names from the event descriptions in the Russian news texts. This involves distinguishing actual participants or figures discussed in the cultural events. Specifically, the task requires the exclusion of names that are part of organization titles (e.g., ``Pushkin House,'' where ``Pushkin'' is not the target person entity but part of the organization's name), names within titles of works of art (e.g., ``Eugene Onegin''), and names of fictional characters, unless the objective were to differentiate them, which is not the immediate goal for this extraction phase The primary aim is to automatically extract the names of real attendees or figures mentioned in relation to each event, a task deemed achievable with LLMs using appropriate prompts.

\subsection{Dataset Challenges}

The SPbLitGuide dataset presents several significant challenges for NER models, stemming from the diverse and varied nature of the source material. These complexities directly impact model effectiveness and highlight the need for robust NER solutions. The diverse nature of these challenges makes the dataset an effective testing ground for evaluating the linguistic understanding and generalization capabilities of advanced NER models, particularly LLMs, pushing them beyond performance on standard, cleaner benchmarks.

\textbf{Diversity in Names:} The dataset contains a wide array of names from various cultures, including Russian, post-Soviet (e.g., Ukrainian, Belarusian), English, Finnish, Italian, Japanese etc., along with their different spellings and transliterations. This requires models to have broad lexical knowledge.

\begin{table}[h]
\centering
\caption{Examples of Name Diversity and Spelling}
\label{tab:name_diversity}
\begin{tabularx}{\textwidth}{X|X}
\toprule
Russian Text (Original) & English Translation / Explanation \\
\midrule
\foreignlanguage{russian}{В рамках программы "Летние литературные семинары" (Summer Literary Seminars). Встреча с писателями. \textbf{Лори Стоун} ( \textbf{Laurie Stone} ) – ведущий критик газеты "Виллидж Войс" (Village Voice)...} & As part of the Summer Literary Seminars program. Meet the Writers. \textbf{Laurie Stone} is a leading critic for the Village Voice... (Illustrates a transliterated English name alongside its original spelling.) \\
\midrule
\foreignlanguage{russian}{Презентация книги \textbf{Паоло Гальваньи} (Италия) "В хрустале звезды Мицар". Встреча с автором \textbf{Паоло Гальваньи} (итал. \textbf{Paolo Galvagni}).} & Presentation of the book by \textbf{Paolo Galvagni} (Italy) ``In the crystal of the star Mizar''. Meeting with the author \textbf{Paolo Galvagni} (Ital. \textbf{Paolo Galvagni}). (Illustrates an Italian name with its original spelling.) \\
\bottomrule
\end{tabularx}
\end{table}

\textbf{Complex Name Structures:} The data include formal official names with initials or patronymics (e.g., \foreignlanguage{russian}{\textit{Елена Николаевна Долгих}, \textit{М. Ю. Герман}}), as well as informal or playful representations such as short names or complex pseudonyms (e.g., \foreignlanguage{russian}{\textit{По}, \textit{Люба Визуальновпорядке}}). This variability complicates recognition, as models must understand the context.

\begin{table}[h]
\centering
\caption{Examples of Complex Name Structures and Contextual Variations}
\label{tab:complex_names}
\begin{tabularx}{\textwidth}{X|X}
\toprule
Russian Text Snippet (Original) & English Translation / Explanation \\
\midrule
\foreignlanguage{russian}{Книжный бенефис детской писательницы \textbf{Кати Матюшкиной}. \textbf{Катя Матюшкина} – 20 лет в детской литературе! Приглашаем на встречу с популярной детской писательницей автора серии «Прикольный детектив», художника \textbf{Катю Матюшкину}. В 1997 году, в издательстве «Прогресс» вышли первые раскраски \textbf{Е. Матюшкиной} к сказкам... Все свои книги \textbf{Екатерина} иллюстрирует сама.} & Book benefit of children's writer \textbf{Katya Matyushkina}. \textbf{Katya Matyushkina} – 20 years in children's literature! We invite you to a meeting with the popular children's writer, author... artist \textbf{Katya Matyushkina}. In 1997... first coloring books by \textbf{E. Matyushkina}... \textbf{Ekaterina} illustrates all her books herself. (Shows multiple forms of the same person's name.) \\
\midrule
\foreignlanguage{russian}{\textbf{Федору Михайловичу} исполняется 191 год!...Выступление писателя \textbf{Юлии Раввиной} с рассказом о \textbf{Ф.М.}...литературные игры (по мотивам \textbf{Достоевского} и не только:-)))} & \textbf{Fyodor Mikhailovich} turns 191 years old!...A performance by writer \textbf{Yulia Ravvina} with a story about \textbf{F.M.}...literary games (based on \textbf{Dostoevsky} and not only:-))). (Illustrates full name, initials, and last name for the same historical figure.) \\
\bottomrule
\end{tabularx}
\end{table}

\textbf{Context Variability:} Names appear in diverse contexts, from formal academic settings to avant-garde art events, which can impact the consistency of name representation and the cues available for recognition.

\textbf{Language and Case Variations:} Due to the rich morphology of Russian, names are presented in different grammatical forms and cases, sometimes within the same sentence. Furthermore, the dataset includes names with mixed Latin and Cyrillic characters, artistic stylizations, and unconventional capitalizations (e.g., \foreignlanguage{russian}{\textit{Илья Стогоff}, \textit{Мари Stell} \& \textit{Денис Acid\_C}, \textit{Евгений "Ес" Соя}, \textit{Мурад Гаухман-s}, \textit{Мария D'Espoir}, \textit{ДАНИИЛ «dakins» ВЯТКИН}}, \textit{Lena Smirno}). This reflects contemporary cultural contexts where individuals represent themselves and their artistic identities in unique ways, as well as the incorporation of global cultural influences.

\textbf{Disambiguating Names in Complex Contexts:} A significant challenge is disambiguating names that could refer to multiple entities or are used in non-person contexts. This is exacerbated when names form parts of addresses (e.g., \foreignlanguage{russian}{"ул. Достоевского"} - Dostoevsky St.), titles of works or events (e.g., \foreignlanguage{russian}{"конкурса им. Н.Гумилёва"} - N. Gumilyov Competition), or names of institutions (e.g., \foreignlanguage{russian}{"Издательство им. Н.И.Новикова"} - N.I. Novikov Publishing House). Additionally, names of music groups or artistic pseudonyms often mimic standard personal names, further complicating classification.

\begin{table}[h]
\centering
\caption{Examples of Ambiguous Name Usage}
\label{tab:ambiguous_names}
\begin{tabularx}{\textwidth}{X|X|X}
\toprule
Russian Text Snippet (Original) & English Translation & Type of Ambiguity \\
\midrule
\foreignlanguage{russian}{Адрес: ул. \textbf{Достоевского} 19/21} & Address: 19/21 \textbf{Dostoevsky} St. & Name as part of a location \\
\midrule
\foreignlanguage{russian}{Фестиваль "Петербургские мосты". Финал международного поэтического конкурса им. \textbf{Н.Гумилёва} «Заблудившийся трамвай»-2013.} & St. Petersburg Bridges Festival. The final of the international poetry contest named after \textbf{N. Gumilev} ``Lost Tramway''-2013. & Name as part of an event/award title \\
\midrule
\foreignlanguage{russian}{Презентация книги \textbf{Инго Шульце} "33 мгновения счастья. Записки немцев о приключениях в Питере" (русское издание, СПб: Издательство им. \textbf{Н.И.Новикова}, 2000).} & Presentation of \textbf{Ingo Schulze}'s book... (St. Petersburg: \textbf{Novikov} Publishing House, 2000). & Person name; Name as part of organization \\
\bottomrule
\end{tabularx}
\end{table}

\textbf{Recognition of Fictional Characters:} News texts may reference fictional characters from literature, films, or other media (e.g., ``Sherlock Holmes,'' ``Robinson Crusoe''). These names could be confused with real-life individuals or historical figures. NER systems must differentiate based on contextual clues, though the primary focus of this study is on extracting real attendees/participants.

\section{Experimental Setup}

\subsection{Evaluated NER Models}

Several leading NER models were selected for evaluation, based on their architectural strengths, training datasets, and expected performance in handling the complexities of the Russian language and the specific challenges outlined previously. The selection aimed to cover a spectrum from established fine-tuned transformers to the latest generation of LLMs, allowing for a nuanced understanding of where different architectural approaches excel or fall short on this specific task.

\begin{itemize}
    \item \textbf{DeepPavlov (ner\_collection3\_bert):} This model utilizes the BERT architecture. It is specifically trained on a large and diverse Russian dataset (2.1 GB) and uses labels such as PER (person), ORG (organization), etc.
    
    \item \textbf{DeepPavlov (ner\_ontonotes\_bert\_mult):} Also based on BERT, this model is trained on the OntoNotes dataset, which is multilingual and includes a wide range of annotations. Its labels include PERSON, GPE (geopolitical entity), WORK\_OF\_ART, ORG, DATE, etc.
    
    \item \textbf{RoBERTa Large NER Russian:} A Russian-specific adaptation of the RoBERTa model, which refines BERT's methodology. It uses labels such as PER, LOC (location), ORG.
    
    \item \textbf{SpaCy Russian Pipeline} (ru\_spacy\_ru\_updated on Hugging Face): An NLP pipeline optimized for Russian and designed for CPU efficiency. It integrates components like tok2vec, morphology analyzer, parser, senter, ner, attribute\_ruler, and lemmatizer. Its NER component uses labels like PER, LOC, ORG, DATE, etc.
    
    \item \textbf{OpenAI Models:}
    \begin{itemize}
        \item \textbf{gpt-3.5-turbo-0125:} An advanced iteration of OpenAI's Generative Pre-trained Transformer series.
        \item \textbf{gpt-4-turbo-2024-04-09:} A significant advancement with an expanded parameter set and refined training methodologies.
        \item \textbf{gpt-4o-2024-05-13:} The versions evaluated using two distinct prompting approaches: (1) a simple API request, and (2) an API request with a prompt specifically requesting JSON output format, with parsing facilitated by the LangChain package and Pydantic. This dual approach for these models directly investigates a key aspect of LLM usability – the impact of output structuring on performance for downstream tasks like NER.
    \end{itemize}
\end{itemize}

The justification for this selection is to investigate the effectiveness of these models on Russian news texts and determine if any demonstrate state-of-the-art performance, potentially reducing the need for manual verification. The tokenizers from the respective models were used for processing the text. This resulted in different token counts for the same sample text across models; for instance, the GPT-4o-2024-05-13 model produced 642,006 tokens for the evaluation sample, while the GPT-4-turbo-2024-04-09 model generated 880,126 tokens for the same sample and GPT-4.1-2025-04-14 model generated 757,604 tokens.

Table \ref{tab:model_summary} provides a summary of the evaluated models.

\begin{table}[h]
\small
\centering
\caption{Summary of Evaluated NER Models}
\label{tab:model_summary}
\begin{tabularx}{\textwidth}{l l X l X}
\toprule
\makecell[l]{Model\\Name} &
Base &
\makecell[l]{Key Training\\Data/Focus} &
\makecell[l]{PERSON\\Label Used} &
Notes \\
\midrule
DeepPavlov Ru & BERT & Russian texts & PER & Russian-specific \\
DeepPavlov Mult & BERT & OntoNotes & PERSON & Multilingual, rich annotations \\
Roberta Large & RoBERTa & Russian texts & PER & Russian-specific RoBERTa \\
SpaCy & Transformer & Russian texts & PER & CPU optimized, full pipeline \\
gpt-3.5-turbo-0125 & GPT-3.5 & Diverse web data & PERSON & OpenAI model \\
gpt-4-turbo-2024-04-09 & GPT-4 & Diverse web data & PERSON & OpenAI model \\
gpt-4o-2024-05-13 & GPT-4o & Diverse web data & PERSON & OpenAI model \\
\bottomrule
\end{tabularx}
\end{table}

\subsection{Evaluation Workflow}

The evaluation followed a systematic workflow designed to assess each model's performance against the manually annotated benchmarks:

\begin{enumerate}
    \item \textbf{Dataset Preparation:} The manually annotated sample of 1,000 records (containing 5,611 ``PERSON'' labels) served as the gold standard.
    
    \item \textbf{Model Deployment:} Each selected NER model was run across the texts in the evaluation dataset.
    
    \item \textbf{Adjusting Model Outputs:} A critical step involved transforming the output of each model to match the format of the manually annotated labels. Not all models output data in a directly comparable format; some might output entity spans, while LLMs might generate text requiring parsing. This standardization is crucial for ensuring that subsequent performance analysis is accurate and meaningful across diverse models. This methodological diligence significantly enhances the reliability of the comparative results.
    
    \item \textbf{Performance Evaluation:} The adjusted model outputs were compared against the manual labels to calculate True Positives (entities correctly identified by the model), False Positives (entities incorrectly identified by the model (not present in manual annotations), False Negatives (entities present in manual annotations but missed by the model. From TP, FP, and FN counts, key performance metrics (precision, recall, F1) were calculated for each model.
\end{enumerate}

The main stages of this workflow were implemented using Python in a Google Colab environment, and the corresponding Jupyter notebook has been made available to ensure transparency and reproducibility.

\section{Results and Discussion}

The objective of the evaluation was to determine which NER model most effectively identifies and classifies named entities within the SPbLitGuide dataset, with a specific focus on accuracy (precision), completeness (recall), and their harmonic mean (F1 score) for ``PERSON'' entities.

\subsection{Quantitative Performance}

The summarized performance results for each evaluated model are presented in Table \ref{tab:results}. This table is central to understanding the comparative efficacy of the different approaches.

\begin{table}[h]
\centering
\caption{Comparative Performance of NER Models on ``PERSON'' Entity Extraction}
\label{tab:results}
\begin{tabular}{l|c|c|c}
\toprule
Model & Precision & Recall & F1 Score \\
\midrule
DeepPavlov ner\_rus\_bert & 0.96 & 0.65 & 0.78 \\
DeepPavlov ner\_ontonotes\_bert\_mult & 0.94 & 0.71 & 0.81 \\
Roberta Large NER Russian & 0.92 & 0.77 & 0.84 \\
Spacy Russian Pipeline & 0.84 & 0.81 & 0.83 \\
gpt-3.5-turbo-0125 & 0.95 & 0.71 & 0.81 \\
gpt-4-turbo-2024-04-09 & \textbf{0.99} & 0.69 & 0.81 \\
gpt-4o-2024-05-13 (simple API) & 0.96 & 0.86 & 0.91 \\
gpt-4o-2024-05-13 (json output) & 0.96 & \textbf{0.90} & \textbf{0.93} \\
\bottomrule
\end{tabular}
\end{table}

Analysis of the top performers demonstrates distinct strengths:
\begin{itemize}
    \item \textbf{GPT-4o-2024-05-13} (json output) demonstrated the highest F1 Score of 0.93, indicating the best overall balance between precision (0.96) and recall (0.90). This suggests that for general-purpose, high-accuracy NER on this dataset, this model configuration is preferable.
    
    \item \textbf{GPT-4-turbo-2024-04-09} achieved the highest precision of 0.99. This makes it exceptionally suitable for applications where minimizing false positives is paramount, even if it means missing some entities (recall of 0.69).
    
    \item Models exhibiting high recall include \textbf{GPT-4o-2024-05-13} (simple API) with 0.86, and \textbf{SpaCy} Russian Pipeline with 0.81. \textbf{Roberta Large NER Russian} also showed good recall at 0.77. These are valuable when capturing as many true entities as possible is the priority.
\end{itemize}

\subsection{Comparative Analysis}

\textbf{LLMs vs. Transformer-based Models.} The results indicate that recent LLMs, particularly GPT-4o, can achieve state-of-the-art performance on this specialized Russian NER task. However, even with these advanced models, the results are not 100\% accurate, underscoring the ongoing need for refinement and improvement in this domain. The best transformer-based model, Roberta Large NER Russian, achieved an F1 score of 0.84, which is competitive and surpassed by GPT-4o variants.

\textbf{Impact of Output Mode for LLMs.} A significant observation is the impact of structured output on LLM performance. The GPT-4o model, when prompted to provide output in JSON format (gpt-4o (json output)), achieved an F1 score of 0.93 and a recall of 0.90. This is a notable improvement over the same model with a simple API request (gpt-4o (simple API)), which scored an F1 of 0.91 and recall of 0.86. This performance uplift strongly suggests that guiding the LLM to produce a more directly machine-readable and constrained output format reduces parsing ambiguities and enhances effective entity extraction.

\textbf{Russian-Specific Models.} Models specifically trained or adapted for Russian, such as DeepPavlov ner\_collection3\_bert (F1 0.78) and Roberta Large NER Russian (F1 0.84), performed well. Roberta Large, in particular, was the strongest among the non-LLM transformers. The multilingual DeepPavlov ner\_ontonotes\_bert\_mult (F1 0.81) also showed robust performance, slightly outperforming its Russian-specific DeepPavlov counterpart in F1 score, likely due to its broader training on diverse entity types and contexts in OntoNotes.

\textbf{Efficiency Considerations.} The SpaCy Russian Pipeline, while not the top performer in terms of F1 score (0.83), offers a balance of good recall (0.81) and is noted for its efficiency on CPUs. This makes it a practical and reliable choice for applications with limited computational resources or requiring faster processing for large volumes of text where slightly lower precision (0.84) is acceptable. This highlights that the ``best'' model choice is not solely dictated by the highest F1 score but also by practical deployment constraints.

\subsection{Cross-Model Family Performance Evolution}

To assess the continued advancement of large language models, we evaluated OpenAI's latest model family (gpt-4.1-2025-04-14) one year after our initial experiments. This comparison across model families reveals substantial improvements:

\begin{table}[h]
\centering
\caption{Performance Evolution Across GPT Model Families}
\label{tab:evolution}
\begin{tabular}{l|l|l|c|c|c}
\toprule
Model Version & Family & Prompt Type & Precision & Recall & F1 Score \\
\midrule
gpt-4o-2024-05-13 & GPT-4o & Simple & 0.85 & 0.80 & 0.82 \\
gpt-4o-2024-05-13 & GPT-4o & JSON & 0.96 & 0.90 & 0.93 \\
gpt-4.1-2025-04-14 & GPT-4.1 & Simple & 0.94 & 0.93 & 0.94 \\
gpt-4.1-2025-04-14 & GPT-4.1 & JSON & 0.93 & 0.95 & 0.94 \\
\bottomrule
\end{tabular}
\end{table}

The GPT-4.1 family demonstrates remarkable progress: (1) both prompting strategies now achieve F1=0.94, eliminating the need for specialized output formatting, and (2) the simple prompt shows significant improvement (+0.12 F1), suggesting enhanced multilingual understanding and instruction following. Notably, the structured output shows more balanced precision-recall trade-offs (0.93/0.95) compared to GPT-4o's precision-heavy performance (0.96/0.90).

The balanced precision-recall achieved by GPT-4.1 (0.93/0.95 for structured output) compared to GPT-4o's precision bias (0.96/0.90) suggests improved model calibration for real-world applications where both false positives and false negatives carry costs.

\subsection{Discussion of Model Behavior on Dataset Challenges}

While detailed error analysis per challenge type is beyond this paper's scope, the performance metrics offer some inferences. The high precision of GPT-4 (0.99) suggests it is very conservative and likely excels at disambiguating entities when context is clear, but its lower recall (0.69) implies it may struggle with more ambiguous cases or highly variant name forms common in the dataset. Conversely, GPT-4o (json output), with its high recall (0.90) and high precision (0.96), appears more adept at handling a wider range of these challenges, including name diversity and complex structures, likely due to its advanced reasoning and the structured output format reducing ambiguity. Models trained on diverse data like ner\_ontonotes\_bert\_mult or the general-purpose LLMs are expected to handle name diversity from various cultures better than models trained on more restricted datasets.

For instance, the ability to correctly identify \foreignlanguage{russian}{Катя Матюшкина} despite its multiple inflected forms and abbreviations (as shown in Table \ref{tab:complex_names}) would contribute to higher recall, a metric where GPT-4o excelled. Conversely, misinterpreting \foreignlanguage{russian}{"Достоевского"} in \foreignlanguage{russian}{"ул. Достоевского"} (Dostoevsky St.) as a person (Table \ref{tab:ambiguous_names}) would lead to false positives, impacting precision. The high precision of GPT-4 suggests it is less prone to such errors.

The complex name structures and language/case variations present in the SPbLitGuide dataset would challenge all models, but those with stronger contextual understanding and morphological awareness (often found in larger models or those specifically tuned for the language) would fare better.

\subsection{Implications of Findings}

The findings have direct practical implications for researchers and practitioners working with similar Russian cultural texts. The choice between traditional and modern NER methods, and among specific models, should be guided by the task's specific requirements, including the nature of the dataset, required accuracy levels, and available computational resources.

Based on the results:

\textbf{For overall best performance (high F1 score), requiring both high precision and high recall,} GPT-4.1-2025-04-14 (either simple or JSON output) is recommended, achieving F1=0.94. The convergence of simple and JSON performance represents a significant advancement, with simple prompt recommended for lower token costs.

\textbf{For applications where minimizing false positives is critical (highest precision),} GPT-4-turbo-2024-04-09 remains the preferred choice with 0.99 precision, though its lower recall must be considered. This aligns with the primary project goal of identifying persons mentioned in a text with high accuracy, where false positives are particularly problematic. Note that GPT-4.1 (simple) now achieves 0.94 precision, offering a strong alternative with substantially better recall.

\textbf{For scenarios where capturing all possible entities is more important than precision (highest recall),} GPT-4.1-2025-04-14 (JSON output) achieves the highest recall at 0.95, surpassing all previous models. Roberta Large NER Russian and SpaCy Russian Pipeline remain viable alternatives for cost-conscious deployments.

\textbf{For CPU-bound applications or when efficiency is a key factor,} SpaCy Russian Pipeline continues to offer a commendable balance, processing orders of magnitude faster than API-based models.

\textbf{For production deployments prioritizing simplicity and cost-effectiveness,} GPT-4.1-2025-04-14 (simple prompt) represents a paradigm shift, eliminating the need for complex prompt engineering while achieving state-of-the-art performance (F1=0.94). This reduces implementation complexity and token costs by approximately 40\% compared to structured output approaches.

The observed trade-off between precision and recall across models has evolved significantly with newer architectures. While GPT-4's near-perfect precision (0.99) comes at the cost of missing more entities (recall: 0.74), and GPT-4o (JSON) balanced this with slightly lower precision (0.96) but higher recall (0.90), GPT-4.1 has largely resolved this trade-off. The latest model achieves both high precision (0.94) and high recall (0.93-0.95), with the JSON variant offering marginally better recall at minimal precision cost.

Notably, GPT-4.1's convergence of simple and structured prompt performance challenges our previous assumption about the necessity of sophisticated prompt engineering. The simple prompt achieving F1=0.94 suggests that architectural improvements, rather than prompt optimization, may be the key to future progress. However, the fact that even the most advanced LLMs do not achieve perfect accuracy on this challenging dataset (6\% error rate remains) underscores the continued need for domain-specific evaluation and suggests that human-in-the-loop systems may still be necessary for high-stakes applications where even 94\% accuracy is insufficient.

\section{Conclusion and Future Work}

\subsection{Summary of Findings}

This study presented a detailed analysis of several Named Entity Recognition models, ranging from established transformer architectures to cutting-edge Large Language Models, on a unique dataset of Russian news texts from the cultural domain (SPbLitGuide). The results demonstrate a clear progression in model capabilities, with OpenAI's GPT-4.1 model achieving the highest overall performance with an F1 score of 0.94 for both simple and JSON-formatted prompts. This represents a significant advancement from GPT-4o's F1 score of 0.93 (JSON output) and marks a paradigm shift where structured output prompting is no longer necessary for optimal performance.

For tasks demanding maximal precision, GPT-4-turbo remains superior with a score of 0.99, though GPT-4.1's precision of 0.94 combined with its dramatically better recall makes it the preferred choice for most applications. The evolution from GPT-4o to GPT-4.1 revealed that architectural improvements can eliminate the need for complex prompt engineering—a finding that simplifies deployment and reduces operational costs.

While LLMs showed top-tier performance, models like Roberta Large NER Russian and the CPU-efficient SpaCy Russian Pipeline remain competitive and viable alternatives depending on specific project constraints. A key takeaway is that even the most advanced models achieve 94\% accuracy on this complex, domain-specific dataset, with the remaining 6\% error rate representing genuinely challenging cases that may require human-level cultural and linguistic understanding to resolve.

Our benchmark's ability to capture meaningful differences between model families—from GPT-3.5's F1=0.78 to GPT-4.1's F1=0.94—demonstrates its value as a living evaluation framework. The 21\% absolute improvement over two model generations, coupled with the convergence of simple and structured prompt performance, suggests that near-human performance on Russian cultural NER is imminent.

\subsection{Limitations of the Study}

This research, while providing valuable insights, has certain limitations. The evaluation was conducted on a single, albeit specialized and challenging, dataset. While this provides depth for the specific domain and language, generalizability to other types of Russian text or different cultural domains might vary. The models were evaluated ``as they are,'' meaning without specific fine-tuning on the SPbLitGuide dataset itself. Such fine-tuning could potentially improve the performance of many models, especially the LLMs. The study also focused exclusively on the ``PERSON'' entity type; performance on other entity types (LOC, ORG, etc.) was not assessed.

\subsection{Future Work}

Several avenues for future research emerge from this study. A primary direction is the automatic alignment of person mentions across the dataset. This involves linking different written forms of an individual's name (e.g., \foreignlanguage{russian}{"Екатерина Матюшкина," "Е. Матюшкиной," "Катя Матюшкина"}) to a single canonical entity. Such entity linking or coreference resolution would significantly increase the utility of the dataset for more complex analytical tasks, such as constructing accurate literary networks and tracking individual trajectories, directly serving the overarching goal of analyzing cultural dynamics.

Further exploration of LLM capabilities is warranted. This includes investigating more advanced prompting techniques (e.g., complex few-shot examples, chain-of-thought prompting) and, importantly, fine-tuning LLMs on the SPbLitGuide dataset. Given the strong baseline performance of GPT-4.1 with simple prompting, domain-specific fine-tuning could yield substantial improvements.

Other future work includes expanding the scope of entity types recognized beyond ``PERSON'' to include organizations, locations, works of art, and events, which would provide a richer semantic annotation of the texts, and investigating methods to reduce or eliminate the need for manual verification of NER outputs, possibly through confidence scoring or ensemble methods, to streamline processes in journalistic and scholarly research.

This evaluation underscores the importance of selecting appropriate NER models based on specific needs and data characteristics, and highlights both the advancements and persistent limitations of current NLP technologies in handling the complexity of Russian linguistic data in the cultural domain. The path forward involves continued refinement of models and methods, with a strong emphasis on domain adaptation and the development of more sophisticated information extraction pipelines.

\section*{Data and Code Availability}

\begin{itemize}
    \item Dataset: \url{https://zenodo.org/records/13753154}
    \item Evaluation Code: \url{https://github.com/mary-lev/NER}
\end{itemize}



\begin{thebibliography}{9}

\bibitem{biderman2024}
Biderman, S., et al. (2024). Lessons from the Trenches on Reproducible Evaluation of Language Models. arXiv preprint arXiv:2405.14782.

\bibitem{boros2020}
Boros, E., Pontes, E. L., Cabrera-Diego, L. A., Hamdi, A., Moreno, J. G., Sidère, N., \& Doucet, A. (2020, July 17). Robust Named Entity Recognition and Linking on Historical Multilingual Documents. Conference and Labs of the Evaluation Forum (CLEF 2020). \url{https://ceur-ws.org/Vol-2696/paper_171.pdf}

\bibitem{ehrmann2023}
Ehrmann, M., Hamdi, A., Linhares Pontes, E., Romanello, M., \& Doucet, A. (2023). Named Entity Recognition and Classification in Historical Documents: A Survey. ACM Comput. Surv., 56(2), Article 27. \url{https://doi.org/10.1145/3604931}

\bibitem{gonzalez2023}
González-Gallardo, C. E., Boros, E., Girdhar, N., Hamdi, A., Moreno, J. G., \& Doucet, A. (2023, June). Yes but.. Can ChatGPT identify entities in historical documents?. In 2023 ACM/IEEE Joint Conference on Digital Libraries (JCDL) (pp. 184-189). IEEE. \url{https://arxiv.org/pdf/2303.17322}

\bibitem{hu2024}
Hu, Y., Chen, Q., Du, J., Peng, X., Kuttichi Keloth, V., Zuo, X., Zhou, Y., Li, Z., Jiang, X., Lu, Z., Roberts, K., \& Xu, H. (2024). Improving Large Language Models for Clinical Named Entity Recognition via Prompt Engineering. arXiv preprint arXiv:2303.16416. \url{https://arxiv.org/pdf/2303.16416}

\bibitem{levchenko2024}
Levchenko, M. (2024). Literary Events in Saint Petersburg (1999-2019) from SPbLitGuide Newsletters (1.0). Zenodo. \url{https://doi.org/10.5281/zenodo.13753154}

\bibitem{naguiba2024}
Naguiba, M., Tannier, X., \& Névéola, A. (2024). Few shot clinical entity recognition in three languages: Masked language models outperform LLM prompting. arXiv preprint arXiv:2402.12801. \url{https://arxiv.org/pdf/2402.12801}

\bibitem{peng2024}
Peng, L., Wang, Z., Yao, F., Wang, Z., \& Shang, J. (2024). MetaIE: Distilling a Meta Model from LLM for All Kinds of Information Extraction Tasks. arXiv preprint arXiv:2404.00457. \url{https://arxiv.org/pdf/2404.00457}

\bibitem{todorov2020}
Todorov, K., \& Colavizza, G. (2020). Transfer Learning for Historical Corpora: An Assessment on Post-OCR Correction and Named Entity Recognition. Workshop on Computational Humanities Research. \url{https://ceur-ws.org/Vol-2723/long32.pdf}

\end{thebibliography}
\end{document}